\documentclass[a4paper,conference]{IEEEtran}

\newcommand*{\rom}[1]{\uppercase\expandafter{\romannumeral #1\relax}}

\newcommand{\subhead}[1]{ 
	\vspace{0.2em}
	 \textbf{{\smash{#1}}}:\xspace}

\newcommand{\ag}{anti-forensic generator\xspace}

\newcommand{\gi}{GAN-generated image\xspace}

\usepackage{times}
\usepackage{epsfig}
\usepackage{graphicx}
\usepackage{amsmath}
\usepackage{amssymb}
\usepackage{comment}
\usepackage{xspace}

\ifCLASSINFOpdf
\else
\fi

\begin{document}
%
\title{Making Generated Images Hard To Spot: A Transferable Attack On Synthetic  Image Detectors}

\author{
\IEEEauthorblockN{Xinwei Zhao}
\IEEEauthorblockA{Department of Electrical and Computer Engineering\\
Drexel Univerisity \\
Philadelphia,  PA,  19104,  USA\\
Email: xz355@drexel.edu}
\and
\IEEEauthorblockN{Matthew C. Stamm}
\IEEEauthorblockA{Department of Electrical and Computer Engineering\\
Drexel Univerisity \\
Philadelphia,  PA,  19104,  USA\\
Email: mcs382@drexel.edu}}

\maketitle

\begin{abstract}
Visually realistic GAN-generated images have recently emerged as an important misinformation threat.  Research has shown that these synthetic images contain forensic traces that are readily identifiable by forensic detectors.
Unfortunately, these detectors are built upon neural networks, which are vulnerable to recently developed adversarial attacks.
In this paper,  we propose  a new anti-forensic attack capable of fooling GAN-generated image detectors.  Our attack uses an adversarially trained generator to synthesize  traces that these 
detectors associate with real images.  
Furthermore, we propose a technique to train our attack so that it can achieve transferability, i.e. it can fool unknown CNNs that it was not explicitly trained against.  
We evaluate our attack through an extensive set of experiments, where we show that our attack can fool eight state-of-the-art detection CNNs with synthetic images created using seven different GANs,  and outperform other alternative attacks.

\end{abstract}

\IEEEpeerreviewmaketitle

\section{Introduction}

Recent technological advances have enabled the creation of synthetic images that are visually realistic.  Generative adversarial networks (GANs)~\cite{GAN2014} in particular have driven this development.  
Several GANs have been proposed that are capable of synthetically generating images of both objects and human faces that are convincingly real to human observers~\cite{starganv2choi2020, stylegan2Karras2019, stylegankarras2019style,  bigganbrock2018large,gauganpark2019gaugan,progankarras2017progressive,  choi2018stargan}. 
Unfortunately, these synthetic image generation techniques can be used for malicious purposes, such as the creation of fake personas to be used as part of misinformation campaigns.

To combat this threat, researchers have developed many techniques to detect GAN-generated images~\cite{cozzolino2018forensictransfer,  mccloskey2018detecting,  nataraj2019detecting, zhang2019detecting, wang2020cnn}  
and to attribute them 
to the specific GAN used to create them~\cite{marra2019, 9035099, 9010964}.
%
%
%
At the same time, adversarial examples have arisen as a new threat to classifiers built from neural networks~\cite{L-BFGS,fgsm,iterative-fg,jsma,cw,pgd}.  
These represent important threats to the forensic community because they can be used as an anti-forensic attack against forensic detectors~\cite{barni2018adversarial,carlini2020evading,counter2017CNN}.   
Recent work from the forensic community,  however, suggests that these attacks may not achieve transferability,  i.e. they may be unable to attack classifiers other than those  they were directly trained against~\cite{transferability_Barni,zhao2020effect}.  


For an anti-forensic attack to be successful,  it must (1)~fool a victim classifier and (2) maintain high visual quality within the attacked image.  Furthermore, it is highly desirable for an attack to (3) transfer to victim classifiers not 
seen  during training and (4) be easily deployable in practical scenarios, i.e. it should work on images of any size, not require specific knowledge of the image window analyzed by a forensic CNN, deploy quickly and efficiently, etc.

In this paper, we propose a new attack that is capable of fooling forensic synthetic image detectors into thinking that {\gi}s are in fact real images.  
This attack achieves each of the four goals described above,  including a significant degree of transferability,  which enables it to attack victim classifiers that are unseen during training.  
Instead of crafting adversarial examples that exploit flaws in forensic detctors,  our attack uses an \ag to synthesize forensic traces associated with real images.  We propose GAN-based approaches for training our \ag for both white-box scenarios and zero-knowledge scenarios.  
Once the \ag is trained,  it can be used to attack images of arbitrary size without requiring re-training or additional tuning to the image under attack.

The main contributions of this work are as follows: 
\vspace{-0.2em}
\begin{itemize}
\setlength\itemsep{-0.0em}

\item We propose a new generative anti-forensic attack that is able to fool CNN-based synthetic image detctors.  Our attack operates by synthesizing forensic traces associated with real images while introducing no perceptible distortions into an attacked image.

\item We propose an ensemble loss training strategy that enables our attack to achieve transferability in zero-knowledge scenarios.  

\item We demonstrate the effectiveness of our attack against many state-of-the-art forensic CNNs,  using synthetic images from a wide variety of different GANs.   

\item We show that our proposed attack achieves in a higher attack success rate, image quality, and transferability than  other alternative attacks, including other adversarial example and GAN-based attacks.

\end{itemize}

\section{Related Work}

Here we briefly review related work on detecting {\gi}s and adversarial attacks. 

%

\subhead{GAN-Generated Image Detectors}
To defend against the misinformation threat posed by synthetic media,  significant research has been done to create \gi detection algorithms~\cite{9010964,  8397040,  cozzolino2018forensictransfer,  mccloskey2018detecting,  nataraj2019detecting, zhang2019detecting, wang2020cnn}.   Previous  research has shown that GANs leave behind forensic traces that are  distinguishable from real images.  These forensic traces left by GANs can be utilized to detect GAN-generated images.  Some approaches operate in a data driven manner~\cite{8397040,  wang2020cnn, cozzolino2018forensictransfer}, while other approaches utilize semantic information~\cite{mccloskey2018detecting} 
or hand-crafted features~\cite{nataraj2019detecting}.  Additionally,  forensic techniques are  also developed to  identify which GAN was used to generate an 
image~\cite{marra2019,  9010964, zhang2019detecting}.  


\subhead{Adversarial Attacks on Forensic Classifiers}
At the same time, adversarial attacks on neural networks have emerged as an important threat~\cite{fgsm,yuan2019adversarial}.  
These attacks can be adapted to attack forensic classifiers~\cite{barni2018adversarial}.   
Roughly speaking, we can group these attacks into two different families: adversarial-example-based attacks and GAN-based attacks.


Adversarial example attacks operate by creating additive image perturbations that cause a victim classifier to misclassify the image.  
Several techniques have been proposed to create these perturbations, including L-BFGS~\cite{L-BFGS},  FGSM and iterative-FG~\cite{fgsm,iterative-fg},  JSMA~\cite{jsma},  CW~\cite{cw},  PGD~\cite{pgd}.   
Attacks based on adversarial examples have been used to forensic algorithms, including camera model identification algorithms~\cite{counter2017CNN} and deepfake detectors~\cite{carlini2020evading}.  Research by Barni et. al has shown, however, that adversarial example attacks do not transfer well to other forensic classifiers~\cite{transferability_Barni}.



Previous research has also shown that GANs can be utilized to construct  
attacks that falsify forensic traces.  
GANs were used by Chen et al. to falsify camera model fingerprints~\cite{mislgan} and by Cozzolino et al. to falsify device fingerprints~\cite{cozzolino2019spoc}.
Kim et.  al used a GAN to remove forensic traces left by median filtering~\cite{residualgan}.  
 However,  research has shown that the  GAN-based anti-forensic attacks also have trouble in transferring~\cite{zhao2020effect}.

\begin{figure}
\centering
\includegraphics[width=0.45\textwidth]{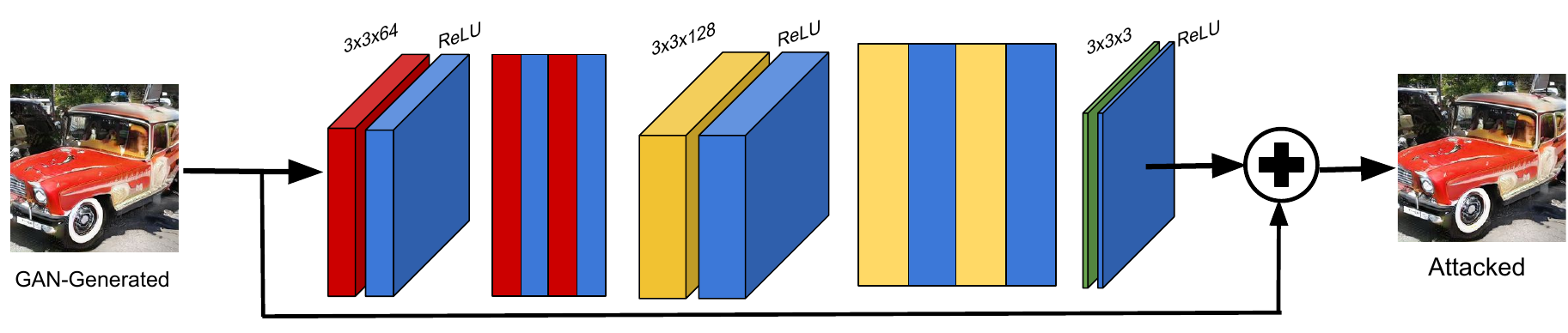}
\caption{Architecture of the proposed anti-forensic generator.}
\vspace{-0.5em}
\label{fig: generator}
\end{figure}

\section{Proposed Attack}


Our attack is designed to modify a \gi $I$ so that a forensic CNN will instead classify it as a `real' image.  
This forensic CNN is alternately referred to as the victim classifier,  and is trained to differentiate between real and {\gi}s.

Our attack operates by passing the  \gi 
through a pre-trained \ag $G$ in order to falsify its forensic traces.  The \ag  is designed to remove forensic traces associated with {\gi}s and synthesize traces associated with `real' images.  
%
As a result, a victim classifier will  classify 
the attacked \gi $G(I)$ as a real one.
Furthermore, the \ag designed to make no changes to the image's contents and to introduce no visually perceptible distortions into the attacked image.  This will prevent a human from visually identifying that an image was attacked.

The \ag in our attack learns to synthesize `real' forensic traces through adversarial training.   
It is trained as part of a GAN in which the discriminator is replaced by a forensic classifier (or set of classifiers) that has been pre-trained to learn the distribution of forensic features associated with real and {\gi}s.  

Different  strategies are used to train the \ag depending on whether the attack is launched in a white-box or zero-knowledge scenario. 
In the white-box scenario, our attack aims to synthesize forensic features with the distribution learned by the victim classifier, even if they deviate from the ideal feature distribution of real images.
%
In the zero-knowledge scenario,  our attack aims to learn the  distribution of forensic features of real images.  However it avoids synthesizing features in regions where different classifiers may 
make different decisions.  
Instead,  it aims to synthesize features that any classifier will likely associate with a real image.

\subsection{Proposed Anti-Forensic Generator Architecture}
The proposed anti-forensic generator consists of a sequence of convolutional layers followed by ReLU activations~\cite{ReLU} shown in Figure~\ref{fig: generator}.   The first three convolutional layers use  64 filters,  the middle three convolutional layers use 128 filters,  and the final convolutional layer uses three filters to reduce the 128 feature maps to a three color channel image.  The output of the generator is the summation of the  input of the generator and the output of the last activated convolutional layer.  The skip connection is designed to give the generator  a better initialization for producing  high visual quality attacked images.
All convolutional layers use $3\times 3$ filter with stride 1.  The small filter size allows the generator to synthesize forensic traces in small areas.  We avoid using any pooling layers  to ensure that the output of the generator is of the same size as the input of the generator.  Therefore,  the proposed anti-forensic generator can be applied to images of arbitrary sizes,  and does not need to be trained for images of different sizes individually.   This characteristic  makes the deployment of the proposed  anti-forensic generator efficient and quick.  

\subsection{Anti-Forensic Generator Training}
When adversarially training the \ag~$G$, we  formulate  a loss function 
to ensure that a attacked image can both fool a victim classifier and maintain high visual quality.  
This  loss function $\mathcal{L}_{G}$ consists of the weighted sum of two terms: the perceptual loss $\mathcal{L}_{p}$ and classification loss $\mathcal{L}_{c}$
\vspace{-0.25em}
\begin{equation}
\label{eq:g_loss}
\mathcal{L}_G=\alpha\mathcal{L}_p + \mathcal{L}_c
\end{equation}
\vspace{-0.25em}
where $\alpha$ is used to balance the trade-off between the visual quality and performance of the attack.

%
%

\subhead{Perceptual Loss} This term is used to minimize distortions introduced by the \ag and control the visual quality of the attacked image.  We define this term as  the mean absolute difference between the \gi $I$ (i.e the input of the generator ) and the attacked image produced by the generaor $G(I)$, such that
\vspace{-0.3em}
\begin{equation}
\mathcal{L}_p = \,\tfrac{1}{N} \|I-G(I) \|_1
\end{equation}
where $N$ is the number of pixels in $I$ and $G(I)$.

\subhead{Classification Loss} This term is used to measure if  the  attacked images produced by the anti-forensic generator can fool the CNN detector used for training.  It allows the generator to learn forensic traces learned by the CNN detector.  The classification loss is provided by the victim  classifier for white-box attacks,   and  is provided by an ensemble of classifiers chosen by the attacker for zero-knowledge attacks.

\subhead{White-Box Attack Training} 
In the white-box scenario,  the attacker has direct access to the forensic CNN under attack.  Hence,  the \ag can be directly trained against the victim classifier.  In this case,  we define the classification
loss $\mathcal{L}_c$ as the softmax cross-entropy between the CNN detector's  output of attacked images and the real class,  
\vspace{-0.0em}
\begin{equation}
	\mathcal{L}_c =-\sum_{k} t_k \log\left(C(G(I))_k\right),
	\label{eq: white-box}
\end{equation}
\vspace{-0.3em}
where $C(\cdot)$ is the victim classifier and  $t_k$ is the $k^{th}$ entry of ideal softmax vector with  a $1$ for the real class and a $0$ for the fake class.  
Defining the classification loss in this manner incentivizes the \ag to learn the victim classifier's model of forensic features from real images.  

\subhead{Zero-Knowledge Attack Training}
In the zero-knowledge scenario, the attacker has no access to the victim classifier that they wish to attack, nor do they know its architecture.  This differs from the black box scenario in which the attacker can probe the victim classifier through an API, then observe the victim classifier's input-output relationship.  Instead, the attacker must rely entirely on the transferability of their attack to fool the victim CNN.

To achieve transferability, we propose adversarially training against an ensemble of forensic classifiers created by the attacker. 
Here,  the classification loss $\mathcal{L}_c$ is formulated as the weights sum of individual classification loss pertaining to each CNN detector in the ensemble,  
\vspace{-0.2em}
\begin{equation}
\mathcal{L}_c= \sum_{s=1}^{S} \beta^{(s)} \mathcal{L}_c^{(s)},
\vspace{-0.1em}
\end{equation}
where $S$ is the number of CNN detectors in the ensemble, $ \mathcal{L}_c^{(s)}$ corresponds to individual classification loss of the $s^{th}$ CNN detector calculated using Equation~\ref{eq: white-box},  $\beta^{(s)}$ corresponds to the weight of  $s^{th}$ individual classification loss. 

Each classifier in the ensemble learns to partition the forensic feature space into separate regions for real and {\gi}s.  By defining the classifier loss in this fashion,  we incentivize the \ag to synthesize forensic features that lie in the intersection of these regions.  
If a diverse set of classifiers are used to form the ensemble, this intersection will likely lie inside the decision region that other classifiers associate with real images.

\section{Experimental Setup}

\subsection{Datasets}

We created two datasets to evaluate our attacks, each containing both real and GAN-generated images.  The first dataset contains only images of human faces, while the second contains images of non-human objects.

\subhead{Human Face Dataset} This dataset consists of real images and GAN-generated images of human faces.   The GAN-generated images were created using StyleGAN~\cite{stylegankarras2019style},  StyleGAN2~\cite{stylegan2Karras2019},  and StarGAN-v2~\cite{starganv2choi2020}.  StyleGAN and StyleGAN2 generated images were downloaded from publicly available datasets shared by Nvidia Research Lab~\cite{stylegandatabase,  stylegan2database}. The StarGAN-v2 generated images were created using pre-trained StarGAN-v2 generator shared at~\cite{starganv2database}.   The real images were downloaded from FFHQ dataset~\cite{stylegan2Karras2019} and CelebA-HQ dataset~\cite{progankarras2017progressive}.  
In total,  the human face dataset contains $66,000$ real images with $44,000$ from FFHQ and $22,000$ from CelebA-HQ; $126,000$ GAN-generated images drawn  equally from  StyleGAN,  StyleGAN2,  and  StarGAN-v2.  

 %

Next,  we partitioned the data into two disjoint training sets, the  \textit{D-set} and the \textit{A-set},  as well as an evaluation set  \textit{Eval-set}.   The \textit{D-set} was used to train the victim forensic CNN detectors.  It contains  $60,000$ GAN-generated images  drawn equally from StyleGAN,  StyleGAN2 and StarGAN-v2; and $60,000$ real images with $40,000$ from FFHQ and $20,000$ from CelebA-HQ.  The \textit{A-set} was used to train the proposed attack.  Since training the proposed attack only requires GAN-generated images,  \textit{A-set} contains $60,000$ GAN-generated images drawn equally from StyleGAN,  StyleGAN2 and StarGAN-v2.

We  benchmarked the baseline performance of the victim forensic CNN detectors and evaluated the performance of our proposed attack against these CNNs using a common evaluation set, \textit{Eval-set}.  The \textit{Eval-set} contains $6,000$ GAN-generated images drawn equally from StyleGAN,  StyleGAN2 and StarGAN-v2; and $6,000$ real images with  $4,000$ from FFHQ and $2,000$ from CelebA-HQ. 
   

\subhead{Object Dataset}This dataset contains real images and GAN-generated images of objects.  The object dataset is a subset of publicly available ForenSynths dataset~\cite{wang2020cnn}.   The ForenSynths dataset was created to demonstrate that CNNs could learn general forensic traces of synthesized images.  Therefore CNNs trained on generated images produced by one GAN method can detect generated images produced by other generative models.  The training set of ForenSynths dataset contains only ProGAN generated images of objects and real images from LSUN dataset~\cite{yu2016lsun}. The testing set of ForenSynths dataset contains varying numbers of generated images produced by  different generative methods.  

From the training set of the ForenSynths dataset,  we created two disjoint training sets:   \textit{D-set} for training victim CNN detectors and \textit{A-set} for training the proposed attack.  \textit{D-set} contains $50,000$ randomly selected ProGAN generated images and $50,000$ randomly selected real images.  
\textit{A-set} contains  $50,000$ randomly selected ProGAN generated images.  

We benchmarked the baseline performance of victim forensic CNN detectors and evaluated the performance of our proposed attack against these CNNs using a common evaluation set,  \textit{Eval-set}.  \textit{Eval-set} contains $4,000$ real images from LSUN,  and all the generated images of objects created by six different GAN methods from the testing set of ForenSynths dataset,  which consists of 26,300 images comprised of 4,000 images from ProGAN~\cite{progankarras2017progressive}, 1,300 from CycleGAN~\cite{cycleganzhu2017unpaired}, 6,000 from StyleGAN~\cite{stylegankarras2019style},  8,000 from StyleGAN2~\cite{stylegan2Karras2019},  2,000 from BigGAN~\cite{bigganbrock2018large},  and 5,000 from GauGAN~\cite{gauganpark2019gaugan}.



%

\subsection{Victim Forensic CNN Detectors}
Before training and evaluating the proposed attack,  we  trained and benchmarked the performance of forensic CNNs on detecting GAN-generated images.  These forensic CNNs were trained as  binary classifiers  to differentiate between real images and GAN-generated images.  
We used eight state-of-the-art CNN models trained as {\gi} detectors, 
 Xception~\cite{xception},  ResNet-50~\cite{resnet},  DenseNet~\cite{densenet},   MISLNet~\cite{mislnet},  PHNet~\cite{phnet},  SRNet~\cite{srnet},  Image CNN~\cite{transfernet},  and CamID CNN~\cite{tuama}.   We trained each forensic CNN individually using the \textit{D-set} of the human face dataset and the object dataset.  This yielded 16 CNN detectors in total and formed the set of victim classifiers we attacked in this paper.

The classification accuracies  of victim CNN detectors on  both datasets are shown in Table~\ref{tab: acc}.  On average,   the average classification accuracy is $95.94\%$ on the human face dataset and $99.14\%$  on the object dataset.   

\begin{table}
\centering
\caption{Classification accuracies of  victim CNN detectors achieved on the human face dataset and  the object dataset.}
\resizebox{0.4\textwidth}{!}{
\begin{tabular}{l|c||c}
\hline
\textbf{CNNs} &\textbf{ Human Face Dataset}&  \textbf{Object Dataset}\\
\hline
\textbf{Xception} &99.67\%&99.75\% \\
\textbf{ResNet-50}& 78.26\%&99.14\%\\
\textbf{DenseNet} &96.39\%&97.04\%\\
\textbf{MISLNet}& 99.93\%&99.15\%\\
\textbf{PHNet} & 95.15\%&99.73\%\\
\textbf{SRNet} &99.50\% &99.78\%\\
\textbf{Image CNN}& 99.49\% & 99.10\%\\
\textbf{CamID CNN} & 99.16\%& 99.46\%\\
\textbf{Avg.}&\textbf{95.94\%} & \textbf{99.14\%}\\
\hline
\end{tabular}}
\vspace{-0.7em}
\label{tab: acc}
\end{table}

\subsection{White-Box Attack}
The first set of experiments was designed to evaluate the effectiveness of our proposed attack in the white-box scenario.  Here,  we assume the attacker has access to the victim classifier and can train directly against it.


For  each victim CNN detector trained on the  human face dataset and the object dataset,   we trained an individual anti-forensic generator to attack it.  To evaluate the performance of the anti-forensic generator,  we used the generator to attack each GAN-generated image in the \textit{Eval-set},   saving the attacked images to  disk as PNG files.   This is to ensure the pixel values of  attacked images reside in the range from 0 to 255.   Next,  we calculated the attack success rate  by using the victim CNN detector  to classify the attacked images.  To evaluate the visual quality of attacked images,   we calculated the mean PSNR between the GAN-generated images and the attacked images. 

The anti-forensic generators presented in this paper were trained from scratch for 32 epochs with a learning rate of $0.0001$.  
Weights were initialized using Xavier initializer~\cite{xavier} and biases were initialized as $0$'s, and were optimized using stochastic gradient descent. 
%
To balance the image  quality and attack success rates,  $\alpha$ in equation~\ref{eq:g_loss} was chosen after a grid search range from $1$ to $200$ with an increment of $20$.  As a result,  $\alpha$ equals to $20$ 
for attacks on the human face dataset and $100$ for attacks on the object dataset. 

\subsection{Zero-Knowledge Attack}
This set of experiments was conducted to evaluate the proposed attack in the zero-knowledge scenario.  We assume the attacker has no knowledge about the victim CNN detector that the investigator would use to classify images.  Particularly,  the attacker has no access to the victim CNN detector and cannot observe any input or output of the CNN detector,  since the investigator may use a private CNN detector that the attacker by all means cannot have access to.   This is a more realistic yet challenging scenario. 
In the zero-knowledge scenario,   we evaluated the transferability of the proposed attack to attack unseen CNN detectors.  

To achieve the transferability,  we built an ensemble of forensic CNN detectors to train the proposed anti-forensic generator.   To mimic the zero-knowledge scenario,   each ensemble used to train the 
attack did not include 
the victim classifier.  
 We assume that if the attacker has no knowledge of the 
 victim classifier's architecture,  
the attacker will use  all available CNNs in their ensemble to strengthen their attack.   
Due to limited GPU memory,  Xception was excluded from the training ensemble in our experiments. 
We trained zero-knowledge attacks using same hyperparameters and other settings as white-box attacks.

\section{Experimental Results and Discussion}

\subsection{White-Box Attack Results}

\label{sec:whiteBoxResults}


\def\tablesize{0.85}
\def\tablesizetwo{1.0}

\begin{table*}[!t]
\label{tab: whitebox}
\centering
\caption{Mean attack success rates and mean image quality achieved for white-box attacks.}
\vspace{-0.5em}
\resizebox{\tablesize\textwidth}{!}{
\begin{tabular}{|l|cc|cc|cc|cc|cc|cc|}
\hline
\multicolumn{13}{|c|}{\textbf{Human Face Dataset}}\\
\hline
 & \multicolumn{2}{|c|}{ \textbf{Residual Gen.}~\cite{residualgan} } & \multicolumn{2}{|c|}{ \textbf{MISLGAN}~\cite{mislgan} } & \multicolumn{2}{|c|}{ \textbf{CW}~\cite{cw}} & \multicolumn{2}{|c|}{ \textbf{FGSM}~\cite{fgsm} }  & \multicolumn{2}{|c|}{ \textbf{PGD}~\cite{pgd}}  & \multicolumn{2}{|c|}{ \textbf{Proposed}} \\\hline
\textbf{CNNs}& \textbf{ASR\%} & \textbf{PSNR}&\textbf{ASR\%}&\textbf{PSNR}& \textbf{ASR\%} & \textbf{PSNR}&\textbf{ASR\%}&\textbf{PSNR}&\textbf{ASR\%}&\textbf{PSNR}&\textbf{ASR\%}&\textbf{PSNR}\\
\textbf{Xception}  & 98.43 &29.01 & 66.68  & 32.24&98.48 &47.29 &54.17 &44.20&100.00&47.32 & 100.00&45.52\\
\textbf{ResNet-50}  & 42.73& 28.82& 66.05 &34.64 &40.91 &38.78 &  37.50& 40.01 &57.20  & 35.13&81.23 &35.97\\
\textbf{DenseNet}  & 99.37 &29.72 & 99.97 & 35.11 &94.32&50.36 &87.50& 44.16& 98.86&51.31 &97.08 &54.28\\
\textbf{MISLNet}  & 98.18 &29.82 & 97.93 &34.50& 95.46&50.53 & 52.65 & 44.26 &100.00&51.17&99.40&51.28 \\
\textbf{PHNet}  & 96.83 &30.21 & 99.68 & 34.84 &98.86&51.05 & 97.73&51.14 &99.24& 51.22&94.94&41.85\\
\textbf{SRNet}   & 97.53&29.46 & 94.28  &33.09&95.83&49.84 & 81.87&51.14 &100.00&51.22&88.05 &50.97\\
\textbf{ImageCNN}  & 66.67 &29.21 & 80.50 &34.36 &100.00 &49.54 & 0.00&44.47 & 94.70&49.04&94.31&53.64 \\
\textbf{CamID CNN}  & 99.80 &28.89& 42.23 &33.30&94.69 &49.25  &0.38 &44.16 & 92.88&51.15 &94.25& 58.10\\ \hline
\textbf{Avg.}  & 87.44 &29.39& 80.92 &34.01 &89.82&48.33&51.47&45.44 &92.86 &48.44 & \textbf{93.66}&\textbf{48.95}\\
\hline \hline
\multicolumn{13}{|c|}{\textbf{Object Dataset} }\\
\hline
 & \multicolumn{2}{|c|}{ \textbf{Residual Gen.} } & \multicolumn{2}{|c|}{ \textbf{MISLGAN} } & \multicolumn{2}{|c|}{ \textbf{CW} } & \multicolumn{2}{|c|}{ \textbf{FGSM} }  & \multicolumn{2}{|c|}{ \textbf{PGD}}  & \multicolumn{2}{|c|}{ \textbf{Proposed}} \\\hline
\textbf{CNNs}& \textbf{ASR\%} & \textbf{PSNR}&\textbf{ASR\%}&\textbf{PSNR}& \textbf{ASR\%} & \textbf{PSNR}&\textbf{ASR\%}&\textbf{PSNR}&\textbf{ASR\%}&\textbf{PSNR}&\textbf{ASR\%}&\textbf{PSNR}\\
\textbf{Xception} & 99.63&39.73 &94.83&32.09 &66.67&50.31 &74.82 &51.19&67.42&51.15 & 99.52&52.59\\
\textbf{ResNet-50} &94.61&38.51 &88.46& 31.50& 89.85&49.58  &  74.79& 51.16& 86.36& 51.17&95.24&55.41\\
\textbf{DenseNet}  &95.22&37.52 &95.15&31.22 &81.23 &50.80 &79.93 & 51.15& 81.81 &51.16&91.16&50.22\\
\textbf{MISLNet} &83.53&40.49 &99.12&31.75 & 64.02&51.08 & 65.91 & 51.26 &71.21&51.16&92.70&50.97\\
\textbf{PHNet} & 96.46&40.22 & 83.07&31.00 &82.20 &50.93 &  81.82&51.15 &83.71&51.15 & 98.02&51.99\\
\textbf{SRNet}  &82.06&37.36 & 94.06&31.68 &61.35 & 46.61& 0.00 &51.16 &72.35&51.15  &82.87&56.41\\
\textbf{ImageCNN}  & 63.42&38.66 &81.09 & 33.43 &67.05 &50.95 &69.70  &51.46& 69.32&51.15 &80.57&55.32 \\
\textbf{CamID CNN}  &80.39&39.02 &75.69& 31.37 &60.61&49.01&59.09&51.34 & 57.20& 51.15&93.05&52.77\\ \hline
\textbf{Avg.}  &86.95&38.94 &88.93&31.76 &71.62&49.91 &63.19&51.23&73.67&51.16&\textbf{91.64}&\textbf{53.21}\\
\hline
\end{tabular}
\label{whiteboxTable}
}
\end{table*}




\begin{table*}[!t]
\label{tab: zero}
\centering
\caption{Mean attack success rates and mean image quality achieved for  zero-knowledge attacks.}
\vspace{-0.5em}
\resizebox{\tablesizetwo\textwidth}{!}{
\begin{tabular}{|l|cc|cc|cc|cc||cc|cc|cc|cc|}
\hline 
 & \multicolumn{8}{|c||}{\textbf{Human Face Dataset}} & \multicolumn{8}{|c|}{\textbf{Object Datset}}  \\
\hline
 & \multicolumn{2}{|c|}{\textbf{CW}} & \multicolumn{2}{|c|}{\textbf{FGSM}}  & \multicolumn{2}{|c|}{\textbf{PGD}}  & \multicolumn{2}{|c||}{ \textbf{Proposed}} & \multicolumn{2}{|c|}{\textbf{CW}}  & \multicolumn{2}{|c|}{\textbf{FGSM}}  & \multicolumn{2}{|c|}{\textbf{PGD}} & \multicolumn{2}{|c|}{ \textbf{Proposed}}
  \\\hline
\textbf{CNNs}& \textbf{ASR\%} & \textbf{PSNR}&\textbf{ASR\%}&\textbf{PSNR}&\textbf{ASR\%}&\textbf{PSNR}&\textbf{ASR\%}&\textbf{PSNR}
& \textbf{ASR\%} & \textbf{PSNR}&\textbf{ASR\%}&\textbf{PSNR}&\textbf{ASR\%}&\textbf{PSNR}&\textbf{ASR\%}&\textbf{PSNR}\\
\textbf{Xception}  & 19.75 &39.58&17.58 &40.07&0.54&38.89& 91.07 &37.93  & 45.70 &40.95&41.67 &40.12 &82.19 &39.06& 96.63&39.95\\
\textbf{ResNet-50}  &19.53 &39.99  &  20.18 & 40.07 & 18.13 &38.15&24.35 & 42.90 &70.20 & 41.09& 50.43 & 40.13& 54.60 & 39.06 & 80.45& 41.75\\
\textbf{DenseNet}   &12.72 &39.97 &4.06 &40.07 &4.87&38.92& 87.30&38.79 &68.74&41.17&52.71& 40.13& 60.87 &38.34 &95.19&40.46\\
\textbf{MISLNet}  &0.50 &39.91 & 0.00 & 40.06 &0.05&38.89&39.80 &38.14 &43.37 &41.63 & 52.92& 40.11 &64.88&39.06&25.04 &41.39\\
\textbf{PHNet} &4.22&40.18 & 0.08&40.07 &14.61& 38.98&91.62&41.69&51.46 &41.38  & 41.67& 40.13 &52.16& 39.09&97.48&42.18\\
\textbf{SRNet}    & 11.91&  39.72&16.45 &40.07&1.08&38.98&88.78 &40.16&11.73 &41.25& 16.45& 40.13 &0.00&39.09&8.49& 39.95\\
\textbf{ImageCNN}  &11.09&39.55 &  11.69&40.03& 14.77&38.87&86.73&41.27&39.34&41.20  &  52.97&40.02&75.92&39.06&60.38& 41.03 \\
\textbf{CamID CNN}  &0.05 &39.84  &0.00&40.07& 10.82& 38.92& 91.17 &42.77&52.70&40.82  &53.57&40.09&85.15&39.13& 99.06&43.43\\ \hline
\textbf{Avg.} & 9.97 &39.84&8.85&40.06&6.47&38.84&\textbf{75.10}&\textbf{40.46} & 52.70 &41.19 &45.30&40.11&59.75&38.99&\textbf{70.30}&\textbf{41.27}\\
\hline
\end{tabular}
\label{zeroTable}
}
\vspace{-0.5em}
\end{table*}


Table~\ref{whiteboxTable} shows the  performance achieved by white-box attacks on both datasets.   Results are presented for our proposed attack, as well as other existing GAN-based and adversarial example based attacks.  Here, the attack success rate (ASR) corresponds to the mean ASR achieved over synthetic images from all GAN generators for a particular victim CNN.
Similarly, the PSNR in this table is defined as the mean PSNR between the original and attacked image.




\subhead{Proposed Attack Performance}
\looseness=-1 The performance of our proposed attack is shown along the rightmost columns of Table~\ref{whiteboxTable}.  
%
On average, our proposed attack achieved an attack success rate of $93.66\%$ against detectors trained on the Human Face Dataset while maintaining an average PSNR of $48.95$.  Similarly, our proposed attack achieved an ASR of $91.62\%$ against detectors trained on the Object dataset while maintaining an average PSNR of $53.21$.  
Example images created by the proposed attack are shown in Figure~\ref{fig: object_demo}.  From this figure, we can see that our attack introduces no visually identifiable artifacts.

These results indicate that our attack can successfully fool a wide variety of CNNs with different architectures trained to detect synthetic images.  In each case, our attack maintained a very high image quality, indicating that our attack is not detectable to the human eye.  
%
%
Additionally, these results show that our attack can make synthetic images created by a wide variety of GANs appear to be real.  
We note that for attacks on the object dataset,  our proposed attack was trained  using only ProGAN generated images.  Despite this, our attack can still reliably fool all detectors used in this experiment.  These results show that our attack can be used on synthetic images made by GANs that our attack was not explicitly trained with.

Finally, we note that our proposed attack outperforms all other adversarial and GAN-based attacks.  These results were significantly more pronounced on the Object Dataset, 
where our attack achieved an ASR  that was $17\%$ higher than then best performing adversarial example attack (PGD).  
These results are discussed in greater detail below.

\subhead{Comparison With Adversarial Example Attacks} 
We  compared our proposed attack with three  well-known adversarial example attacks:  Carlini-Wagner (CW)~\cite{cw},  Fast Gradient Sign Method (FGSM)~\cite{fgsm},  and Projected Gradient Descent (PGD)~\cite{pgd}.  
We used the  CleverHans toolbox~\cite{papernot2018cleverhans} to launch these attacks, then saved attacked images as PNGs.  
We note that due to computational limitations,  adversarial example attacks were evaluated using a representative subset of the \textit{Eval-set}.  This is because adversarial example attacks must be individually trained for each image that they attack.  
For these experiments, our \textit{Eval-set} for each adversarial example attack corresponded to 6,240  images for the Human Face Dataset and 12,480 images for the Object Dataset.




From Table~\ref{whiteboxTable}, we can see that at the same image quality our attack outperformed adversarial example attacks.  On the Human Face Dataset, our attack achieved an attack success rate of $93.66\%$.  By contrast, the CW attack achieved an ASR of $89.82\%$,  FGSM achieved  an ASR of $51.47\%$, and PGD achieved an ASR of $92.86\%$.  These results were more pronounced on the Object Dataset.  Our attack achieved an average ASR of $91.64\%$, while CW achieved an ASR of $71.62\%$, FGSM achieved an ASR of $63.19\%$, and PGD achieved an ASR of $73.67\%$.  
These results show that even in white-box scenarios, our attack yields important performance advantages over adversarial example attacks.

We note that in these experiments, we chose the parameters for each adversarial example attack such that the mean PSNRs of  attacked images were similar to those obtained by our attack.  This was done for two reasons.  The first was to maintain a fair comparison between our attack and these attacks.  The second was to maintain acceptable visual quality for anti-forensic applications.  It is well-known that a stronger attack typically means more visible perturbations.  While this can be acceptable in computer vision algorithms, it is not acceptable for anti-forensic applications.  Images with implausible visual distortions such as speckles will  be rejected as inauthentic by humans.  As a result, anti-forensic attacks have a higher visual quality requirement.




\subhead{Comparison With Other Anti-Forensic Generators} 
We also compared  our proposed attack to existing GAN-based anti-forensic attacks~\cite{residualgan,mislgan}. 
%
The results in Table~\ref{whiteboxTable} show that our proposed attack can outperform these attacks in terms of both ASR and image quality.  On the Human Face Dataset, our attack generator achieved an ASR that was $6.22\%$ than the residual generator~\cite{residualgan} and $12.74\%$ higher than MISLGAN~\cite{mislgan} while maintaining at least 15dB PSNR higher in image quality.  
We note that the image qualities achieved by the residual generator and MISLGAN are low enough that visually detectable artifacts and distortion are present.  At comparable image qualities, our attack's performance gains are likely to be significantly more pronounced.   
Similar results were achieved on the Object Dataset, where our attack generator achieved an ASR that was $4.69\%$ than the residual generator~\cite{residualgan} and $12.74\%$ higher than MISLGAN~\cite{mislgan} while maintaining at least 15dB PSNR higher in image quality.  

\def\figheight{29mm}
\begin{figure}[!t]
\includegraphics[height=\figheight]{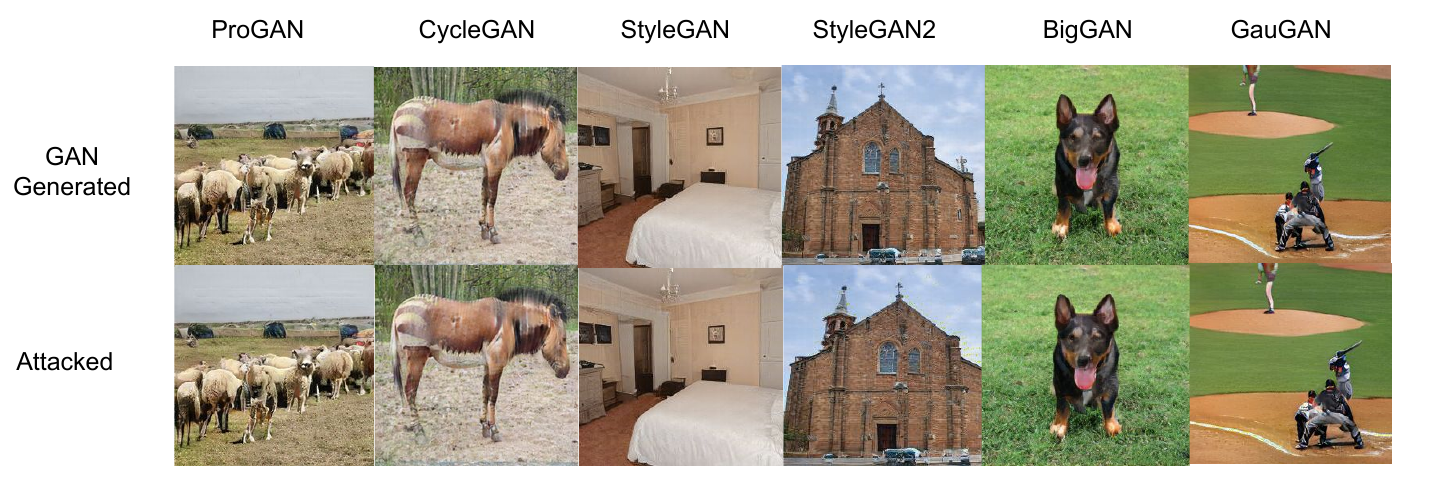}
\vspace{-1em}
\caption{Attacked images produced by the proposed white-box attack.}
\vspace{-0.5em}
\label{fig: object_demo}
\end{figure}

\subsection{Zero-Knowledge Attack Results}


Next, we evaluated attack performance in the zero-knowledge scenario, in which the victim CNN is unknown to the attacker and cannot be probed in a black box manner.  As a result, the attack's success relies entirely on transferability. 

Table~\ref{zeroTable} shows the  performance achieved on both datasets by zero-knowledge attacks.  
Each entry in the table corresponds to the mean attack success rate (ASR) and mean PSNR  achieved by an attack when attacking a particular victim CNN detector.  
The other anti-forensic GANs were omitted from these experiments due to both space limitations, because our anti-forensic generator outperformed these other generators in  white-box results presented in Section~\ref{sec:whiteBoxResults}, and because these generators introduced visible artifacts into attacked images.

\subhead{Proposed Attack Performance}
The results in Table~\ref{zeroTable} show that our proposed attack can achieve significant transferability, resulting in  strong ASRs even in the zero-knowledge scenario.  On the Human Face Dataset, our attack achieved an ASR of $75.10\%$ while maintaining an average PSNR of 40.46.  Similarly, our attack achieved an ASR of $70.30\%$ while maintaining an average PSNR of 41.27.
These results are particularly important because in many realistic conditions, forensic detectors will be kept private and will not be publicly queriable (e.g. detectors used by governmental and defense agencies, law enforcement, etc.).  In these conditions, both white-box and black-box attacks are infeasible.

\begin{figure}[!t]
\centering
\includegraphics[height=\figheight]{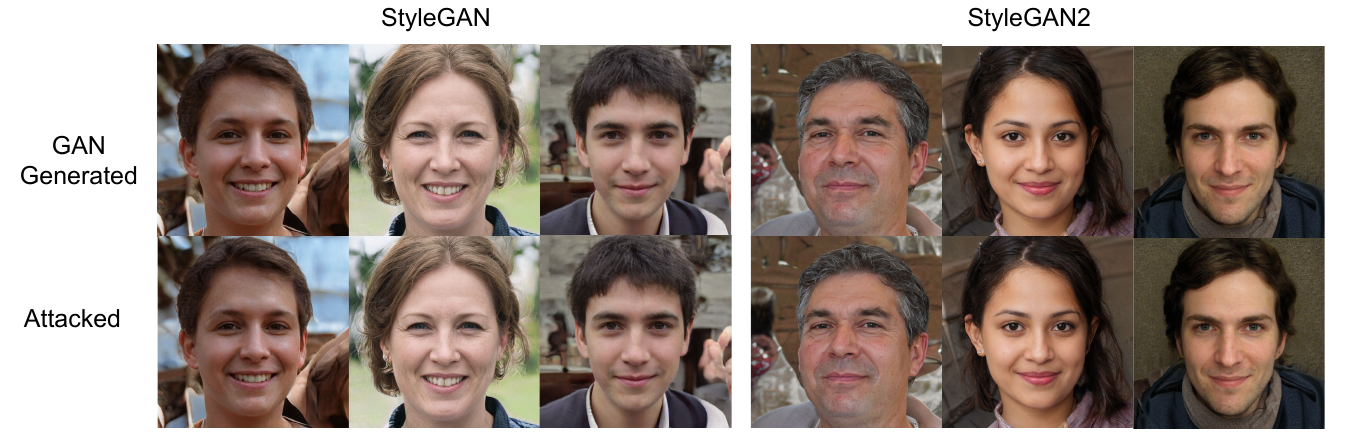}
\vspace{-2em}
\caption{ Attacked images produced by  zero-knowledge attack. }
\vspace{-0.6em}
\label{fig: face_demo}
\end{figure}

Images from the Human Face Dataset attacked using our zero-knowledge attack are shown in Fig.~\ref{fig: face_demo}.  From this figure, we can see that our attack maintains high visual quality without introducing perceptible distortions or artifacts.

We note that when compared to the white-box scenario, the performance drop incurred by our attack in the zero-knowledge scenario was largely attributable to a small number of CNNs.  For the Human Face Dataset our attack had lower transferablity to both MISLnet and ResNet-50, while on the Object Dataset, our attack had lower transferability to both MISLnet and SRNet.  
Excluding these CNN, our average ASR is $89.45\%$ on the Human Face Dataset and is $88.20\%$ on the Object Dataset.  
We note that adversarial example attacks had difficulty attacking these CNNs too.  Future studies of this may provide insight into designing CNN architectures that are more resilient to anti-forensic attacks.

\subhead{Comparison with Adversarial Example Attacks} The zero-knowledge performance of the CW,  FGSM and PGD adversarial example attacks are also shown in Table~\ref{zeroTable}.
 To measure the transferability of an adversarial example attack against a particular victim CNN detector,  we launched the attack against every other CNN detectors,  then used the victim CNN detector to classify the attacked images. 
 For a fair comparison,  we chose the parameters for each adversarial example attack such that the mean PSNRs of  attacked images were comparable to PSNRs achieved by our attack.
 

From Table~\ref{zeroTable}, we can see that adversarial example attacks were broadly unsuccessful on the Human Face Dataset.  The most successful attack was the CW attack, achieving an ASR of $9.97\%$.  Adversarial example attacks were more successful on the Object Dataset, but still achieved ASRs significantly lower than our proposed attack.

\section{Conclusion}
In this paper, we proposed a new attack to fool GAN-generated image detectors.  Our attack uses an adversarially trained generater to synthesize forensic traces that these detectors associate with `real' images.
We proposed training protocols to produce both white-box as well as zero-knowledge attacks.  The latter protocol, which is based on training against an ensemble of classifiers, enables our attack to achieve transferability to unseen victim classifiers.  Through a series of experiments, we demonstrated that our attack does not create perceptible distortions in attacked images, and can fool eight different \gi detectors.  Furthermore,   the proposed attack outperforms other alternative attacks in both white-box and zero-knowledge scenarios.

\bibliographystyle{IEEEabrv}
\bibliography{refs/citations}

%


\end{document}